\documentclass[lettersize,journal]{IEEEtran}
\usepackage{amsmath,amsfonts}
\usepackage{algorithmic}
\usepackage{algorithm}
\usepackage{array}
\usepackage[caption=false,font=normalsize,labelfont=sf,textfont=sf]{subfig}
\usepackage{textcomp}
\usepackage{stfloats}
\usepackage{url}
\usepackage{verbatim}
\usepackage{graphicx}
\usepackage{cite}
\usepackage{xcolor}
\usepackage{times}
\usepackage{epsfig}
\usepackage{amsmath}
\usepackage{amssymb}
\usepackage{multirow}
\usepackage{makecell}
\usepackage{lineno}
\usepackage{tabularx}
\usepackage{relsize}
\usepackage{xmpmulti}
\usepackage[accsupp]{axessibility}
\usepackage{multirow, makecell}
\usepackage{arydshln}

\newcommand{\etal}{\textit{et al}.}
\newcommand{\ie}{\textit{i}.\textit{e}.}
\newcommand{\eg}{\textit{e}.\textit{g}.}

\begin{document}

\title{Modality Mixer Exploiting Complementary Information for Multi-modal Action Recognition}

\author{Sumin Lee, Sangmin Woo, Muhammad Adi Nugroho, Changick Kim~\IEEEmembership{Senior,~IEEE,}
\thanks{This work was conducted by Center for Applied Research in Artificial Intelligence (CARAI) grant funded by DAPA and ADD (UD230017TD).}
\thanks{S. Lee, S. Woo, M.A. Nugroho, and C. Kim are with the School of Electrical Engineering, Korea Advanced Institute of Science and Technology (KAIST), Daejeon 34141, Republic of Korea. (e-mail: \{suminlee94, smwoo95, madin, changick\}@kaist.ac.kr)}
}

\markboth{IEEE TRANSACTIONS ON IMAGE PROCESSING,~Vol.~00, No.~0, April~2023}%
{Lee \MakeLowercase{\textit{et al.}}: Modality Mixer Exploiting Complementary Information for Multi-modal Action Recognition}
\IEEEpubid{0000--0000/00\$00.00~\copyright~2021 /IEEE}


\maketitle

\begin{abstract}
Due to the distinctive characteristics of sensors, each modality exhibits unique physical properties.
For this reason, in the context of multi-modal action recognition, it is important to consider not only the overall action content but also the complementary nature of different modalities.
In this paper, we propose a novel network, named Modality Mixer (M-Mixer) network, which effectively leverages and incorporates the complementary information across modalities with the temporal context of actions for action recognition.
A key component of our proposed M-Mixer is the Multi-modal Contextualization Unit (MCU), a simple yet effective recurrent unit.
Our MCU is responsible for temporally encoding a sequence of one modality (\eg, RGB) with action content features of other modalities (\eg, depth and infrared modalities).
This process encourages M-Mixer network to exploit global action content and also to supplement complementary information of other modalities.
Furthermore, to extract appropriate complementary information regarding to the given modality settings, we introduce a new module, named Complementary Feature Extraction Module (CFEM).
CFEM incorporates sepearte learnable query embeddings for each modality, which guide CFEM to extract complementary information and global action content from the other modalities.
As a result, our proposed method outperforms state-of-the-art methods on NTU RGB+D 60, NTU RGB+D 120, and NW-UCLA datasets.
Moreover, through comprehensive ablation studies, we further validate the effectiveness of our proposed method.

\end{abstract}

\begin{IEEEkeywords}
Multi-modality, action recognition, recurrent unit
\end{IEEEkeywords}

\section{Introduction} 
\label{chap:intro}
Humans perceive and interact with their environment through  a combination of sensory inputs including audio, visual, and tactile information.
Due to recent advancements in sensor technology, there has been significant research interest in multi-modal learning within the realm of computer vision.
Toward this direction, for video action recognition, many multi-modal methods have been developed, which achieved higher performance than other methods based on a single modality.
\begin{figure}[t]
    \centering
    \includegraphics[width=0.5\textwidth]{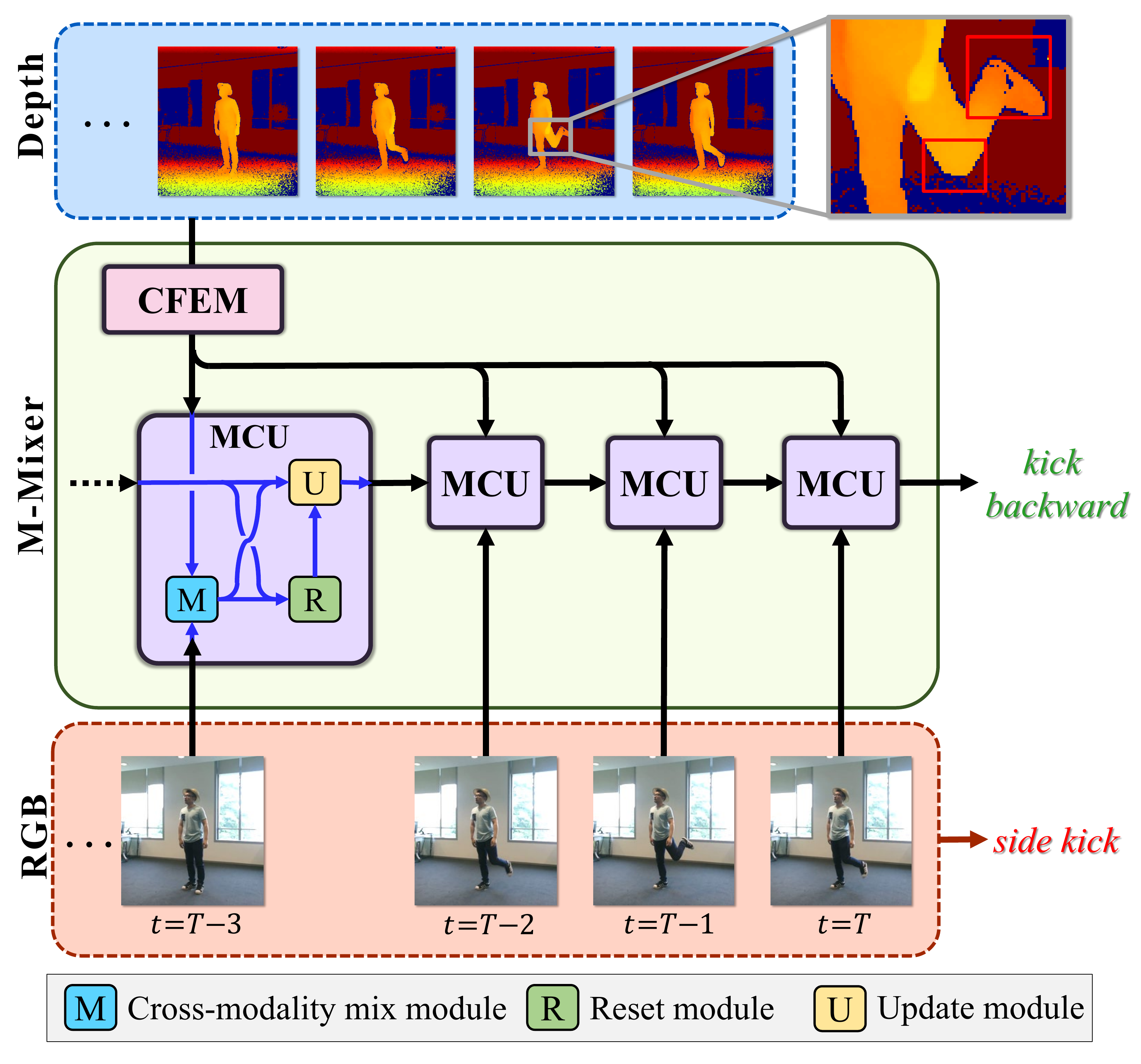}
    \caption{\textbf{Multi-modal Action Recognition with Modality Mixer (M-Mixer) network.}
    When solely relying on the appearance information from the RGB modality, the action `\textit{kick backward}' is prone to being misclassified to `\textit{\textcolor{red}{side kick}}'.
    However, by incorporating the depth modality, which represents the 3D structure of the scene, it becomes possible to capture the foot orientation accurately.
    In the proposed M-Mixer, MCU effectively supplements the RGB information with action content information from the depth frames extracted by Complementary Feature Extraction Module (CFEM).
    By identifying that the foot is going behind the knee, which is represented by a more yellowish in the image (indicating a closer proximity to the camera), M-Mixer correctly classifies the action as `\textit{\textcolor{green}{kick backward}}.'
    Here, although we assume the use of RGB and depth inputs, we depict only RGB stream for clarity.
    }
    \label{fig:intro}
\end{figure}

Earlier studies in action recognition predominantly focused on a single modality, specially RGB videos, and emphasized spaito-temporal modeling~\cite{3d_temp:i3d, 3d_temp:r3d, 3d_temp:nonlocal, 3d_temp:slowfast}.
In recent years, there have been various studies utilizing multiple modalities such as RGB, optical flow, and depth, and exploring appropriate methods for modality fusion~\cite{rgbd_3, mstd, rgbd_4, shahroudy2017deep, liu2018viewpoint, garcia_admd, garcia_eccv, garcia_dmcl, mmar_lstm}.
Due to the different properties of sensors, each modality possesses different key characteristics that contribute to the overall action recognition.
\IEEEpubidadjcol
For example, as illustrated in Fig.~\ref{fig:intro}, the action `kick backward' may be incorrectly predicted as `side kick' when employing only the RGB modality~\cite{ntu120}.
This misclassification occurs due to the difficulty in perceiving the orientation of the left foot.
However, the depth data can indicate that the foot is going behind the knee, leading to the correct action class `kick backward'.
As such, while RGB images provide visual appearance information, depth data conveys the 3D structure of 2D frames, which complements the RGB modality.
Consequently, multi-modal action recognition requires considering two crucial factors: 1) complementary information across modalities and 2) temporal context of action.


In this paper, to address these factors, we propose a novel network, called Modality Mixer (M-Mixer).
The proposed M-Mixer consists of three key parts: 1) extraction of complementary information from other modalities, 2) temporally encoding video frame features with complementary features, and 3) fusion of modality features.
By taking feature sequences from multiple modalities as inputs, our M-Mixer temporally encodes each feature sequence with action content features from other modalities, which are called a cross-modal action content.
The cross-modal action content includes modality-specific information and the overall activity of videos.
To consolidate the encoded features from each modality, we employ a multi-modal feature bank. 
The multi-modal feature bank combines the encoded modality-specific features by incorporating and enhancing multi-modal action information.
The final score is obtained by performing the read operation on the multi-modal feature bank.

We also introduce a simple yet effective recurrent unit, called Multi-modal Contextualization Unit (MCU), which plays a vital role in the M-Mixer network.
The proposed MCU performs temporal encoding of the given modality sequence, while augmenting it with complementary information of other modalities.
Because each MCU is dedicated to a specific modality, we describe our MCU in detail from an RGB perspective, as illustrated in Fig.~\ref{fig:intro}.
MCU consists of three modules: cross-modality mix module, reset module, and update module.
Concretely, given an RGB feature at certain timestep and a cross-modality action content feature, the cross-modality mix module models their relationship and supplements complementary information to the RGB feature. 
Some existing works~\cite{hori2017attention,liu2023multi} employ similarity-based attention for modality fusion.
However, these methods can not fully leverage complementarity between modalities due to their inherent heterogeneity~\cite{baltruvsaitis2018multimodal, wang2022deep}.
On the other hand, the proposed method combines the given modality feature and the cross-modality action content feature using a linear transformation to calculate weights.
The two features are integrated by weighted-summation.
Then, reset and update modules learn the relationships between the integrated feature of the current timestep and the previous hidden state feature.
With these modules, our MCU exploits complementary information across modalities and global action content during temporal encoding. 

In order to exploit suitable complementary information, we introduce a new module, named Complementary Feature Extraction Module (CFEM).
This module addresses the variability in required complementary information depending on the modality setting.
Even from the same modality, different modalities may require different types of complementary information.
To accommodate this variability, CFEM incorporates separate learnable query embeddings for each modality.
The query embeddings are trained to extract complementary information and global action content from other modalities, which are relevant to the designated modality.
By combining MCU and CFEM, our M-Mixer network is able to assimilate richer and more discriminative information from multi-modal sequences for action recognition.
It is worth noting that our M-Mixer is not limited to only two modalities and can be extended to incorporate more modalities.

We extensively evaluate our proposed method on three benchmark datasets (\eg, NTU RGB+D 60~\cite{ntu60}, NTU RGB+D 120~\cite{ntu120}, and Northwestern-UCLA (NW-UCLA)~\cite{nwucla}).
Our M-Mixer network achieves the state-of-the-art performance of 92.54\%, 91.54\%, and 94.86\% on NTU RGB+D 60, NTU RGB+D 120, and NW-UCLA datasets with RGB and depth modalities, respectively.
Furthermore, using RGB, depth, and infrared modalities, the M-Mixer network achieves superior performance compared to previous approaches on NTU RGB+D 60 and NTU RGB+D 120 datasets, with accuracies of 93.16\% and 92.66\%, respectively.
Through comprehensive ablation experiments, we validate the effectiveness of our proposed method.

Our main contributions are summarized as follows:
\renewcommand\labelitemi{\tiny$\bullet$}
\begin{itemize}
\item We investigate how to take two important factors into account for multi-modal action recognition: 1) complementary information across modality, and 2) temporal context of an action.
\item We propose a novel network, named M-Mixer, with a new recurrent unit, called MCU.
By effectively modeling the relation between a sequence of one modality and action contents of other modalities, our MCU facilitates M-Mixer to exploit rich and discriminative features.
\item Furthermore, we introduce a complementary feature extraction module (CFEM). Each query embedding in CFEM, corresponding to specific modality, allows for addressing the variability in required complementary information based on the specific modality setting.
\item To fuse the modality feature encoded by MCU, we employ a multi-modal feature bank. The multi-modal feature bank captures a multi-modal action feature by incorporating action content information across modalities and time.
\item We achieve state-of-the-art performance on three benchmark datasets. Moreover, we demonstrate the effectiveness of the proposed method through comprehensive ablation studies.

\end{itemize}

This paper is an extended version of our previous conference paper~\cite{m-mixer} that investigated the effectiveness of exploiting complementary information during temporal encoding.
Compared with our earlier work, we enhance the method by employing CFEM to learn complementary information based on the modality combination and utilizing a multi-modal feature bank for effective feature fusion.
Also, through more extensive experiments, we validate the effectiveness of our proposed method in multi-modal action recognition.

\section{Related Work}
\label{chap:rw}
Video action recognition has emerged as a prominent task in the field of video understanding.
Over the past decade, video action recognition has made remarkable advancements due to the rise of deep learning and the accessibility of extensive video datasets.
In the early stage, deep learing models adopted a two-stream structure capturing appearance and motion data separately~\cite{2stream:2stream, 2stream:fusion, 2stream:tsn}.
However, due to the computational cost of optical flow, other approaches focused on learning motion features solely from RGB sequences~\cite{rgb3d:d3d, rgb3d:dance, rgb3d:fixed, rgb3d:flow, rgb3d:guided, rgb3d:mars} 
For example, Stroud \etal~\cite{rgb3d:d3d} and Crasto \etal~\cite{rgb3d:mars} suggested learning algorithms that distill knowledge from the temporal stream to the spatial stream in order to reduce the two-stream architecture into a single-stream model.
After then, 3D convolution networks~\cite{3d_temp:i3d, 3d_temp:r3d, 3d_temp:slowfast, 3d_temp:x3d} were proposed, leading to significant performance improvements.
Notably, SlowFast network~\cite{3d_temp:slowfast} employs two pathways to handle two different frame rates and capture spatial semantics and motion.
To address the limitations of conventional CNN in the large-range dependencies, some methods are proposed to capture long-term spatio-temporal representations~\cite{3d_temp:nonlocal,qiu2019learning,varol2017long}.
Recently, due to the huge success of transformers in image domain, transformer-based approaches~\cite{atn,vivit, vidswin, motionformer, mtv} have been introduced and achieved performance.

With the advancement of sensor technologies, action recognition in multi-modal setting has attracted research interest~\cite{wang2020makes, alayrac2020self, caesar2020nuscenes,audio2,audio1, tu2019action}.
While sensor-based methods utilizing gyroscopes and accelerometers~\cite{koo2022contrastive, duhme2022fusion, mondal2020new} have been explored, there is also ongoing research on approaches that combine RGB and skeleton information~\cite{skel_1, skel_2, cui2022pose}.

Among the various modalities, RGB and depth are commonly used in combination~\cite{rgbd_1, rgbd_2, rgbd_3, rgbd_4}.
Shahroudy \etal~\cite{shahroudy2017deep} proposed a shared-specific feature factorization network based on autoencoder structure for RGB and depth inputs.
Liu \etal~\cite{liu2018viewpoint} introduced a method of learning action features that are insensitive to camera viewpoint variation.
Wang~\cite{rgbd_aaai} proposed a Convolutional neural Network (c-ConvNet) that enhances the discriminative information of RGB and depth modalities.
Dhiman \etal~\cite{mstd} introduced a two-stream view-invariant framework with motion stream and shape temporal dynamics (STD) stream for RGB and depth modalities.
In~\cite{garcia_eccv, garcia_admd}, Garcia \etal explored the frameworks involving distillation and privileged information; although these methods are trained with both RGB and depth data, a hallucination network of depth enables classifying actions with only RGB data.
Garcia \etal~\cite{garcia_dmcl} introduced an ensemble of three specialist networks for RGB, depth, and optical flow videos, called the Distillation Multiple Choice Learning (DMCL) network. 
In DMCL, three specialist networks collaboratively strengthen each other through late fusion.
Wang \etal~\cite{mmar_lstm} proposed a hybrid network based on CNN (\eg, ResNet50~\cite{resnet} and 3D convolution) and RNN (\eg, ConvLSTM~\cite{convlstm}) to fuse RGB, depth, and optical flow modalities.
Woo \etal~\cite{actionmae} explored robust fusion techniques for multi-modal action recognition, and proposed a modular network, called ActionMAE.

\begin{figure*}[t]
    \centering
    \includegraphics[width=\textwidth]{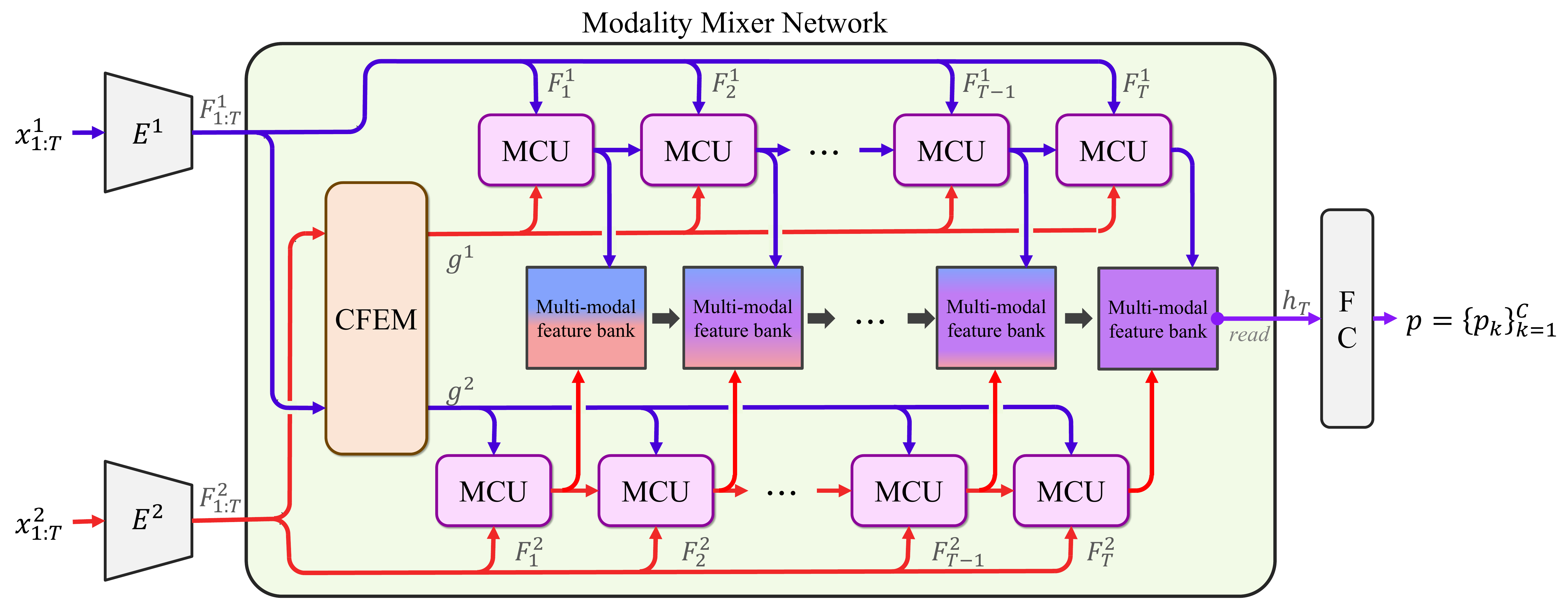}
    \caption{\textbf{Modality Mixer network.} 
    We illustrate an example of using two modalities in this figure. 
    M-Mixer network consists of Complementary Feature Extraction Module (CFEM), Multi-modal Contextualization Unit (MCU) and a multi-modal feature bank.
    The M-Mixer takes a feature sequence $F^i_{1:T}$ as input, obtained from a frame sequence $x^i_{1:T}$ through a feature extractor $E^i$.
    In the first step, CFEM calculates a cross-modal action content feature $g^i$ based on $F^j_{1:T}$, where $j \neq i$.
    In this example, $F^2_{1:T}$ is utilized to compute $g^1$, while $F^1_{1:T}$ is used to calculate $g^2$.
    Then, our MCU encodes the temporal information of a feature sequence of $i$-th modality $f^{i}_{1:T}$,incorporating the cross-modal action content feature $g^{i}$.
    By comparing $f^{i}_{1:T}$ with $g^{i}$ during temporal encoding, MCU takes into account both complementary information across modalities and overall action contents of a video.
    To consolidate multi-modal action information across modalities, the multi-modal feature bank accumulates the hidden state features from all modalities.
    Finally, the probability distribution over $C$ action classes is computed with the final hidden state feature $h_T$, which is read from the multi-modal feature bank.
    Through this process, the proposed M-Mixer network effectively integrates diverse and informative details from multi-modal sequences, enhancing its capability for accurate action recognition.
    In the illustrated figure, the \textcolor{blue}{blue} and \textcolor{red}{red} lines indicate the streams of modality 1 and 2, respectively, and the \textcolor{violet}{purple} line represents the fusion of modalities.}
    \label{fig:overview}
\end{figure*}

Many studies have emphasized the importance of learning complementary information in multi-modal tasks~\cite{wang2022deep, baltruvsaitis2018multimodal, ntu120}.
In this paper, we focus on extracting and exploiting complementary information.
To this end, we propose a novel network, called, Multi-modal Mixer (M-Mixer) network, with Contextualization Unit (MCU), and Complementary Feature Extraction Module (CFEM).
By encoding a feature sequence with cross-modality content features from other modalities, our M-Mixer network enables the exploration of complementary information across modalities and temporal action content.
Note that our M-Mixer network is not limited to specific types and the number of video modalities, making it versatile and adaptable to various scenarios.

\section{Proposed Method}
\label{chap:mixer}
In this section, we first describe the overall architecture of our proposed Modality Mixer (M-Mixer) network and then explain the proposed Modality Contextualization Unit (MCU) and Complementary Feature Extraction Module (CFEM) in detail.
In Fig.~\ref{fig:overview}, the framework of our M-Mixer network is illustrated, assuming the use of two modalities.

\subsection{Modality Mixer Network}
The goal of our M-Mixer network is to generate rich and discriminative features for action recognition of videos with $N$ different modalities.

Given a video of length $T$ for the $i$-th modality, a feature extractor $E^i$ converts a sequence of frames, $x^i_{1:T}\in \mathbb{R}^{3\times T\times H\times W}$, to a sequence of frame features, $ \bar{\mathbf{F}}^i_{1:T} \in \mathbb{R}^{d_f \times T \times h \times w}$, as follows:  
\begin{align}
    \bar{\mathbf{F}}^i_{1:T} &= E^i \left (x^i_{1:T} \right ),
\end{align}
where $H$ and $W$ denote the height and width of an input frame, $h$ and $w$ are the height and width of $\bar{\mathbf{F}}$, and $i = 1,2,\cdots, N$.
To adjust the dimensionality, we apply frame-wise 2D convolution with kernel size of $1\times 1$ to $\bar{\mathbf{F}}$, resulting in $\mathbf{F}^i_{1:T} \in \mathbb{R}^{d_h \times T \times h \times w}$.
Then, the proposed M-Mixer network takes the extracted feature sequences $\mathbf{F}^i_{1:T}$ as inputs.

In M-Mixer, CFEM first generates a set of cross-modality action content features $G = \left\{ g^1, g^2, ..., g^N \right\}$ from $\mathbf{F}_{1:T}$, as follows:
\begin{align}
    g^1, g^2, ..., g^N &= \textrm{CFEM} \left (\mathbf{F}_{1:T}^{1}, \cdots , \mathbf{F}_{1:T}^{N} \right ),
\end{align}
where  $g^i \in \mathbb{R}^{d_h}$.
The proposed CFEM leverages complementary information from other modalities to enhance the representations of each modality.
A learnable query embedding of each modality in CFEM facilitates the extraction of appropriate complementary information for different combinations of modalities.
The details of CFEM are explained in Sec.~\ref{cfem}.

Our M-Mixer network contains $N$ MCUs with each MCU dedicated to a specific modality
Before being input into an MCU, average pooling is applied along the spatial dimension to $\mathbf{F}_{1:T}$ to obtain a feature sequence $f_{1:T}^i \in \mathbb{R}^{d_h \times T}$.
MCU encodes $f_{1:T}^i$ with the cross-modality action content feature $g^i$ in a temporal manner and generates a hidden state  $h^i_t\in \mathbb{R}^{d_h}$ as follows:
\begin{align}
    h^i_t = \,&\texttt{MCU}^i \left (f^i_t, g^i\right ),
\end{align}
where $\texttt{MCU}^i$ denotes an MCU for the $i$-th modality.
By augmenting $g^i$, our MCU exploits complementary information as well as global action content.

To fuse the encoded hidden state features of modalities, we employ a multi-modal feature bank.
The multi-modal feature bank $\mathbf{M}$ has $K$ location vectors with size of $d_h$ (\ie, $\mathbf{M} \in \mathbb{R}^{K \times d_h}$). 
At each step, the location vectors capture and integrates multi-modal action information from the encoded features of different modalities.
Firstly, the dimensionality of $h^i_t$ is reduced to $\left \lfloor d_h / N \right \rfloor$, as follows:
\begin{align}
    \hat{h}^i_t = \mathbf{W}_{h}^i h^i_t,
\end{align}
where $\mathbf{W}_{h}^i \in \mathbb{R}^{d_h \times \left \lfloor d_h / N \right \rfloor} $ is a learnable parameters for the $i$-th modality. 
Then, the update attention score $\alpha^{(t)}$ is computed by comparing the similarity between the previous feature bank $M^{(t-1)}$ and the hidden state features of all modalities: 
\begin{align}
    \alpha^{(t)} =  \sigma \left( \mathbf{M}^{(t-1)} \left( \underset{\forall i}{\mathlarger{\parallel}} \hat{h}^i_t \right) \right).
\end{align}
Here, $\sigma$ represents the sigmoid function and $\parallel$ indicates vector concatenation.
With $\alpha^{(t)}$, the multi-modal feature bank is updated as follows:
\begin{align}
    \hat{\mathbf{M}}^{(t)} = \alpha^{(t)} \otimes \mathbf{M}^{(t-1)}& + (1-\alpha^{(t)}) \otimes \left( \underset{\forall i}{\mathlarger{\parallel}} \hat{h}^i_t \right),\\
    \mathbf{M}^{(t)} =&  \hat{\mathbf{M}}^{(t)} \mathbf{W}_{u},
\end{align}
where $\mathbf{W}_{u} \in \mathbb{R}^{d_h \times d_h}$ is a trainable matrix and $\otimes$ denotes the element-wise multiplication.
Through iterative updates, the multi-modal feature bank accumulates multi-modal action information.
After the full iteration of $T$ steps, the multi-modal action feature $h_T$ is calculated by read operation from $\mathbf{M}^{(T)}$:
\begin{align}
    h_T = W_{r} \mathbf{M}^{(T)},
\end{align}
where $W_{r} \in \mathbb{R}^{K} $ is a trainable parameter.

To obtain the final probability distribution $p=\{p_c\}_{c=1}^C$ over $C$ action classes, we employ a fully connected layer to process $h^i_T$ for all modalities, as follows:
\begin{align}
    p = \xi \left (\mathbf{W}_p h_T +b_p\right ),
\end{align}
where $\xi$ indicats the softmax function, $p_c$ is a probability of the $c$-th action class, $\mathbf{W}_p \in \mathbb{R}^{d_h \times C}$ is a learnable matrix, and $b_p \in \mathbb{R}^{C}$ is a bias term.

To train our M-Mixer network, we define a loss function $L$ based on the standard cross-entropy loss as follows:
\begin{align}
    L = \sum_{c=1}^{C} y_c\log \left (p_c \right ),
\label{loss}
\end{align}
where $y_c$ is the ground-truth label for the $c$-th action class.

\subsection{Multi-modal Contextualization Unit}
\begin{figure}[t]
    \centering
    \includegraphics[width=0.5\textwidth]{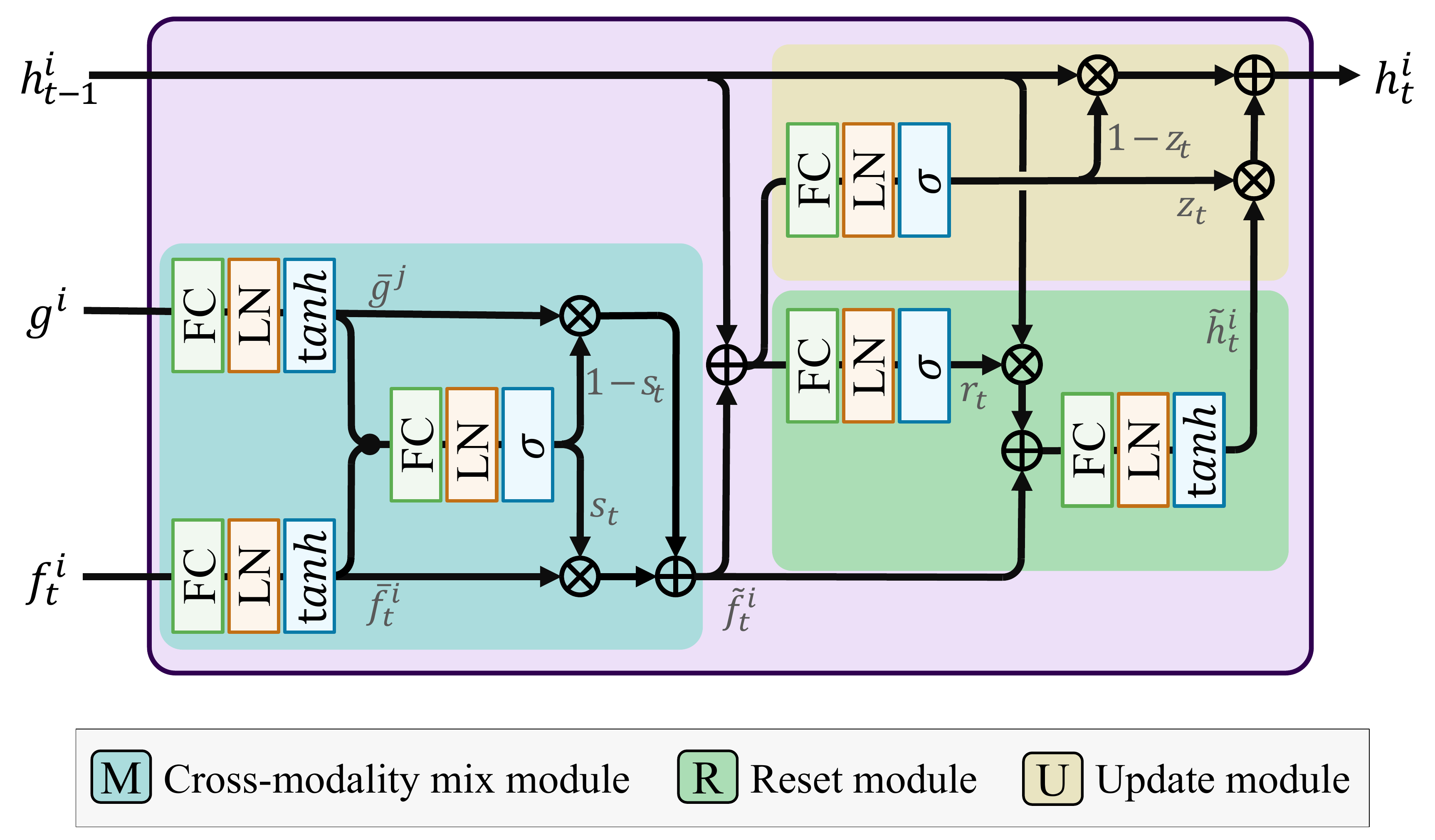}
    \caption{\textbf{Multi-modal Contextualization Unit (MCU).} 
    Our MCU consists of three modules: cross-modality mix module, reset module, and update module. 
    In a cross-modality mix module, a cross-action content $g^i$ is adaptively integrated with $f^i_t$, providing complementary information and the overall action content.
    A reset gate $r_t$ in the reset module serves to distinguish between information to be dropped and information to be taken from previous hidden state $h^i_{t-1}$ and an supplemented feature $\tilde{f}^i_{t}$.
    In an update module, an update gate $z_t$ is computed to update previous hidden state $h^i_{t-1}$.
    }
    \label{fig:mcu}
\end{figure}

We describe our new recurrent unit, MCU, which is the core component of the proposed M-Mixer network.
As described in Fig.~\ref{fig:mcu}, our MCU consists of three submodules: cross-modality mix module, reset module, and update module.
At the $t$-th timestep, the proposed MCU takes $f^i_t$ and $g^i$ to contextualize a modality-specific feature with a cross-modality action content feature.
This strategy enables MCU to supplement with complementary information from other modalities in terms of global action content.
As a result, the proposed MCU exploits rich and well-contextualized features for action recognition.

\subsubsection{Cross-modality Mix Module} First, ${f}^i_t$ and ${g}^i$ are projected to the same embedding space, as follows:
\begin{align}
    \bar{f}^i_t & = \eta \left ( \texttt{LN} \left ( \mathbf{W}_f f^i_t \right ) \right ), \\
    \bar{g}^i &= \eta \left (\texttt{LN} \left ( \mathbf{W}_g g^i \right )\right ),
\end{align}
where $\eta$ represents the tangent hyperbolic function, $\mathbf{W}_g \in \mathbb{R}^{d_h(N-1) \times d_h}$ and $\mathbf{W}_f \in \mathbb{R}^{d_h \times d_h}$ are trainable matrices, and $\texttt{LN}$ denotes the layer normalization.
Note that we exclude a bias term for simplicity.

Next, an integration score $s_t$ is computed to determine how much representations of target modality and other modalities are activated, as follows:
\begin{align}
    s_t &= \sigma \left (\texttt{LN} \left (\mathbf{W}_s [\bar{f}^i_t \parallel \bar{g}^i] \right ) \right ),
\end{align}
where $\mathbf{W}_s \in \mathbb{R}^{d_h \times 2d_h}$ is a weight matrix.
Then, $\bar{f}^i_t$ and $\bar{g}^i$ are combined to the supplemented feature $\tilde{f}^i_t$, as follows:
\begin{align}
    \tilde{f}^i_t& = s_t \otimes \bar{f}^i_t + \left (1-s_t \right ) \otimes \bar{g}^i.
\end{align}

\subsubsection{Reset and Update Module}
Our reset and update modules learn relationships between the supplemented feature $\tilde{f}^i_t$ and previous hidden state $h^i_{t-1}$.
In the reset module, a reset gate $r_t$ effectively drops and takes information from $h^i_{t-1}$ and $\tilde{f}^i_t$.
And the update module measures an update gate $z_t$ to amend previous hidden state $h^i_{t-1}$ to current hidden state $h^i_{t}$.
We compute $r_t$ and $z_t$, as follows:
\begin{align}
    r_t &= \sigma \left ( \texttt{LN} \left ( \mathbf{W}_{hr} \left (  \tilde{f}^i_t + h^i_{t-1} \right ) \right ) \right ),\\
    z_t &= \sigma \left (\texttt{LN} \left (\mathbf{W}_{hz} \left (\tilde{f}_t^i + h^i_{t-1} \right ) \right ) \right ),
\end{align}
where $\mathbf{W}_{hr} \in \mathbb{R}^{d_h \times d_h}$ and $\mathbf{W}_{hz} \in \mathbb{R}^{d_h \times d_h}$ are learnable parameters.
Then, the hidden state $h^i_{t-1}$ is updated with $z_t$, as follows:
\begin{align}
    h^i_t &= z_t \otimes \tilde{h}^i_t +\left (1-z_t \right )\otimes h^i_{t-1},
\end{align}
where $\tilde{h}$ is defined as:
\begin{align}
    \tilde{h}^i_t = \eta \left (\texttt{LN} \left (\mathbf{W}_{hh}\left (r_t \otimes h^i_{t-1} + \tilde{f}^i_t \right ) \right ) \right ).
\end{align}
Here, $\mathbf{W}_{hh} \in \mathbb{R}^{d_h \times d_h}$ is a trainable matrix.

\begin{figure}[t]
    \centering
    \includegraphics[width=0.5\textwidth]{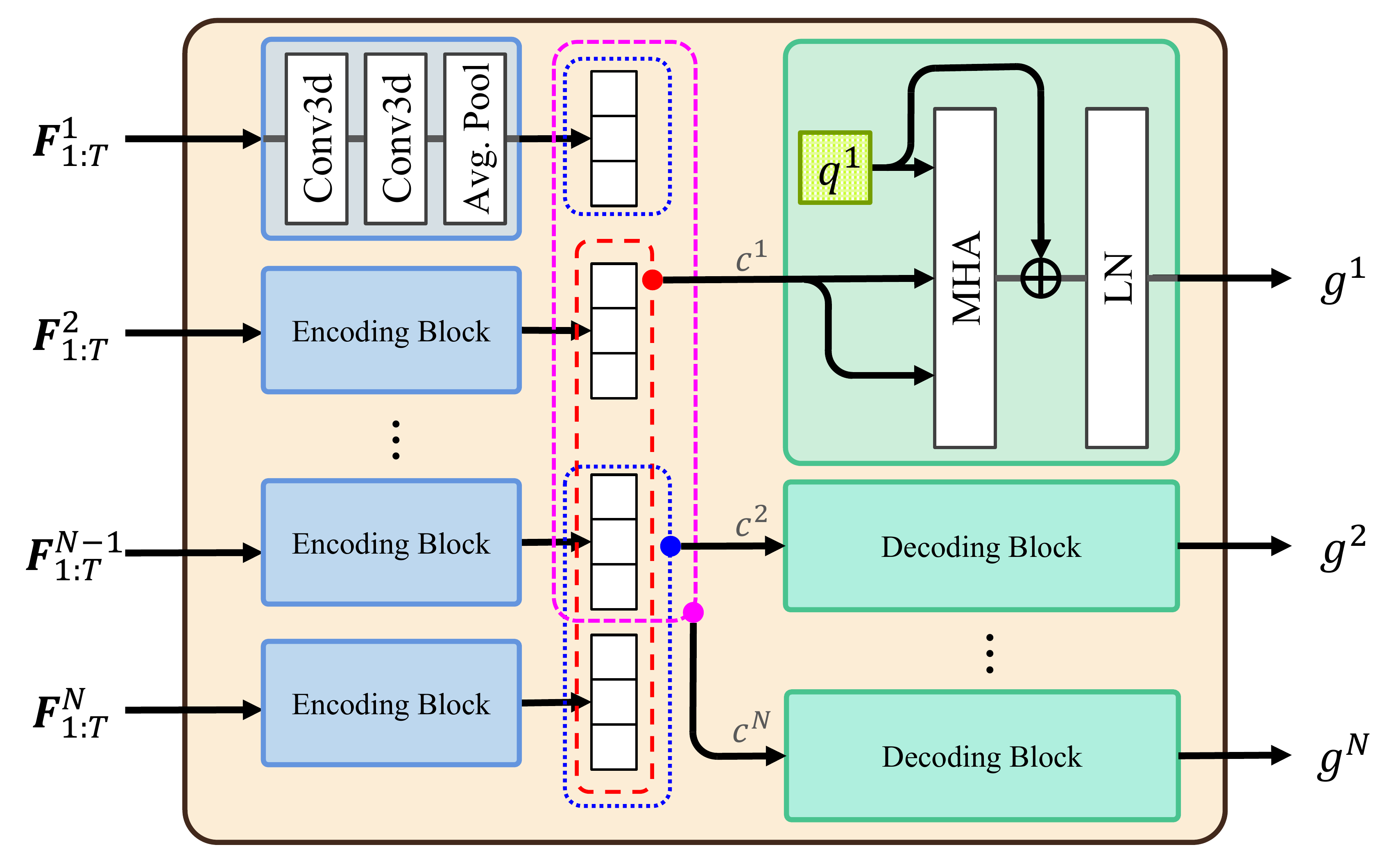}
    \caption{\textbf{Complementary Feature Extraction Module (CFEM).} 
    The proposed CFEM consists of two parts: an encoding block and a decoding block. 
    In the encoding block, the features of each modality are encoded in a spatio-temporal manner by applying 3D convolution layers and average pooling. 
    In the decoding block, a learnable query embedding extracts complementary information and global action content through multi-head attention layer.}
    \label{fig:cfem}
\end{figure}

\subsection{Complementary Feature Extraction Module (CFEM)}
\label{cfem}
As illustrated in Fig.~\ref{fig:cfem}, CFEM consists of two parts: encoding blocks and decoding blocks with each modality having one of each.
The encoding block is composed of two 3D convolution layers with a kernel size of $4 \times 1 \times 1$ and a stride of $2 \times 1 \times 1$ and the average pooling.
By employing 3D convolution layers, the encoding block models the spatio-temporal dependencies between frames in the input feature $F_{1:T}$
And then, global average pooling is applied across the temporal dimension, and the resulting feature is flattened along the spatial axis.
Consequently, each encoding block transforms the input feature $\mathbf{F}^i_{1:T}$ into the aggregated feature $\hat{f}^i \in \mathbb{R}^{d_h \times (h \times w)}$.
We empirically demonstrate that reducing the temporal axis is more effective than reducing the spatial axis or both in Sec.~\ref{sec:ablation_mmmb}.

In the decoding block, the action content feature $c^i$ is computed by concatenating $\hat{f}^j$ from other modalities, as follows:
\begin{align}
    c^i &=  \underset{\forall j}{\mathlarger{\mathlarger{\mathlarger{\parallel}}}} \hat{f}^j, \; \textrm{where} \, j \neq i.
\end{align}
Each decoding block comprises a learnable query embedding $q^i$, a multi-head attention layer, and a layer normalization layer.
The query embedding $q^i$ facilitates the multi-head attention layer to extract complementary features to the $i$-th modality and global action content from $c^i$.
The cross-modality action content feature $g^i$ is derived by $q^i$ via a multi-head attention operation, as follows:
\begin{align}
    g^i &=  \textrm{LN} \left( \textrm{MHA} \left( q^i, c^i + \texttt{pos}, c^i\right) + q^i\right),
\end{align}
where \texttt{pos} is the positional embedding vector of $c^i$, LN indicate layer normalization, and MHA stands for multi-head attention that takes the query, key, and value as inputs.

\section{Experiments}
\label{chap:exp}

\subsection{Dataset}
\subsubsection{NTU RGB+D 60.}NTU RGB+D 60~\cite{ntu60} is a large-scale human action recognition dataset, consisting of 56,880 videos.
It includes 40 subjects performing 60 action classes in 80 different viewpoints.
As suggested in~\cite{ntu60}, we follow the cross-subject evaluation protocol.
For this evaluation, this dataset is split into 40,320 samples for training and 16,560 samples for testing.

\subsubsection{NTU RGB+D 120.} As an extended version of NTU RGB+D 60, NTU RGB+D 120~\cite{ntu120} is one of the large-scale multi-modal dataset for video action recognition.
It contains 114,480 video clips of 106 subjects performing 120 classes from 155 different viewpoints.
We follow the cross-subject evaluation protocol as proposed in~\cite{ntu120}.
For the cross-subject evaluation, the 106 subjects are divided into 53 subjects for training and the remaining 53 subjects for testing.

\subsubsection{Northwestern-UCLA (NW-UCLA).}
NW-UCLA~\cite{nwucla} is composed of 1475 video clips with 10 subjects performing 10 actions.
Each scenario is captured by three Kinect cameras at the same time from three different viewpoints.
As suggested in~\cite{nwucla}, we follow the cross-view evaluation protocol, using two views (View 1 and View 2) for training and the other one (View 3) for testing.

\subsection{Implementation Details}
For the feature extractor for each modality, we use a ResNet-18 or a ResNet34~\cite{resnet} for NTU RGB+D 60~\cite{ntu60} and NTU RGB+D 120~\cite{ntu120} and ResNet-18 for NW-UCLA~\cite{nwucla}, which are initialized with pretrained weights on ImageNet~\cite{imagenet}.
We set the size of the hidden dimension in MCU, $d_h$, to 512 when using two modalities, and to 588 when using three modalities.
The size of multi-modal feature bank $K$ is set to 8.
The input of each modality is a video clip uniformly sampled with temporal stride 8.
For the training procedure, we adopt random cropping and resize each frame to $224 \times 224$. We also apply random horizontal flipping and random color jittering for RGB videos.
We convert the depth and IR frames into color images using the jet colormap as following~\cite{garcia_admd}.

To train our M-Mixer network, we use 4 GPUs of RTX 3090. 
We use the Adam~\cite{adam} optimizer with the initial learning rate of $10^{-4}$.
A batch size per GPU is set to 8 on NTU RGB+D 60 and NTU RGB+D 120.
Due to the small number of training samples, we use a single GPU with batch size 8 on NW-UCLA.

\subsection{Study of Multi-Modal Contextualization Unit (MCU)}
In this section, we investigate how effective the proposed MCU is and examine the importance of cross-modality information during temporal encoding.
All experiments in this section are conducted on NTU RGB+D 60~\cite{ntu60} using RGB and depth modalities with a ResNet18~\cite{resnet} backbone.
To solely see the effects of MCU and cross-modality action contents, we simplify the experimental setup.
Instead of using CFEM and a multi-modal feature bank, we employ average pooling along the spatial-temporal axis and simple concatenation, respectively.
Specifically, the cross-modal action content of $i$-th modality $g^i$  is defined as follows:
\begin{align}
g^i &=  \underset{\forall j}{\mathlarger{\mathlarger{\mathlarger{\parallel}}}} s^j, \; \textrm{where} \, j \neq i.
\label{eq:gi}
\end{align}
Here,  $s^j$ is the spatio-temporally average pooled feature of $F^j_{1:T}$, where $j$ is the index of the modality.

\subsubsection{Comparison with RNNs and Transformer}
\begin{table}[t]
\renewcommand{\arraystretch}{1.1}
\begin{center}
\begin{tabular}{c|c}
\hline
Method & Accuracy (\%) \\ \hline
LSTM~\cite{lstm} & 84.28 \\
LSTM~\cite{lstm} + CM & 85.29\\
GRU~\cite{gru} & 84.87 \\
GRU~\cite{gru} + CM & 86.29\\ \cdashline{1-2}
Transformer & 89.39\\
Transformer + CM & 89.45\\  \cdashline{1-2}
MCU & \textbf{90.77} \\ \hline
\end{tabular}
\end{center}
\caption{\textbf{Comparions with LSTM, GRU, Transformer, and MCU.}
'CM' refers to the utilization of cross-modality action content feature by concatenating it with the input.
The best scores are marked in \textbf{bold}.}
\label{tab:rnn_ablation}
\end{table}

To study the effectiveness of MCU, we conduct ablation experiments by replacing our MCU in M-Mixer with conventional recurrent units (\ie, LSTM~\cite{lstm} or GRU~\cite{gru}) or Transformer.
For each modality, we employ a separate LSTM, GRU, or Transformer and final predictions are calculated by concatenating the output features from all modalities.
Table~\ref{tab:rnn_ablation} presents the performances of the modified M-Mixer network using LSTM, GRU, and Transformer.
While `LSTM' and `GRU' only take a feature sequence $f^i_{1:T}$ as input, `LSTM+CM' and `GRU+CM' utilize the concatenated feature of $f^i_t$ and the cross-modality action content feature $g^i$ as inputs.
For the experiments with Transformer, we use the architecture of a Transformer encoder~\cite{vit} with 4 heads and 2 layers.
The output of a learnable class token is used to predict the action class.
Similar to the recurrent units, `Transformer' indicates the use of only the input feature sequence, while `Transformer+CM' refers to the encoding of the feature sequence along with the cross-modality action content.

From the results of the conventional recurrent units with and without $g^i$, we observe that MCU effectively learns the relations between the current feature and its cross-modality action content to explore discriminative action information.
Thanks to the incorporation of overall action information and complementary information from the cross-modality action content, the proposed MCU achieves performance gains of 6.49\%p and 5.90\%p over LSTM and GRU, respectively.
Also, due to the cross-modality mix module, MCU obtains 5.48\%p and 4.48\%p performance increase over LSTM+CM and GRU+CM, respectively.
The results of the experiments with the Transformer indicate that there is not a significant performance difference between Transformer and Transformer+CM.
This observation is attributed to the fact that the Transformer utilizes a similarity-based attention mechanism.
Despite having 2.4 times more parameters than MCU, the attention mechanism of the Transformer cannot fully capture the complementarity among modalities.
Our MCU achieves performance gains of 1.38\%p and 1.32\%p over Transformer and Transformer+CM, respectively.

\begin{table}[t]
\renewcommand{\arraystretch}{1.1}
\begin{center}
\begin{tabular}{c|ccc|c}
\hline
Exp. & \begin{tabular}[c]{@{}c@{}}Cross-modality\\ mix module\end{tabular} & \begin{tabular}[c]{@{}c@{}}Cross-modality\\ action content \end{tabular} & LN & \begin{tabular}[c]{@{}c@{}}Acc.\\ (\%)\end{tabular} \\ \hline \hline
\uppercase\expandafter{\romannumeral1}& \checkmark &  &  & 86.82\\
\uppercase\expandafter{\romannumeral2}& &  \checkmark&  &  87.76\\
\uppercase\expandafter{\romannumeral3}& \checkmark&  \checkmark&  &  89.97 \\ \cdashline{1-5}
\uppercase\expandafter{\romannumeral4}& \checkmark&  \checkmark&  \checkmark& \textbf{90.77} \\ \hline
\end{tabular}
\end{center}
\caption{\textbf{Ablation study on MCU.}
MCU has three important factors: a cross-modality mix module, cross-modality action content, and layer normalization indicated LN.
The best scores are marked in \textbf{bold}.}
\label{tab:mcu_ablation}
\end{table}

\subsubsection{Ablations of the modules in MCU}
We validate the effects of three important factors in our MCU: the cross-modality mix-module, the cross-modality action content, and the layer normalization.
To observe the abilities of each  factor, we conduct ablation experiments on these three factors and report the performances in Table~\ref{tab:mcu_ablation}.
In Exp. \uppercase\expandafter{\romannumeral1}, we replace the cross-modality action content $g^i$ with the self-modality action content $s^i$ and turn off the layer normalization.
Experiment \uppercase\expandafter{\romannumeral2} aims to investigate the effect of the cross-modality mix module by replacing it with simple concatenation and disabling the layer normalization.
Lastly, in Exp. \uppercase\expandafter{\romannumeral3}, we only turn off the layer normalization in our MCU.
Comparing the results of Exp. \uppercase\expandafter{\romannumeral1} and Exp. \uppercase\expandafter{\romannumeral3}, it is evident that utilizing cross-modality action content leads to an accuracy improvement of 3.15\%p, reaching 89.97\%.
In the following section, Sec~\ref{sec:cm-action-content}, we delve into a detailed analysis of the efficacy of cross-modality action content.
By comparing Exp. \uppercase\expandafter{\romannumeral2} and Exp. \uppercase\expandafter{\romannumeral3}, we observe that using the cross-modality mix module improves the performance from 87.76\% to 89.97\%.
Finally, we obtain the best performance of 90.77\% with all three components in Exp. \uppercase\expandafter{\romannumeral4}.

\begin{figure*}[t]
    \centering
    \includegraphics[width=\textwidth]{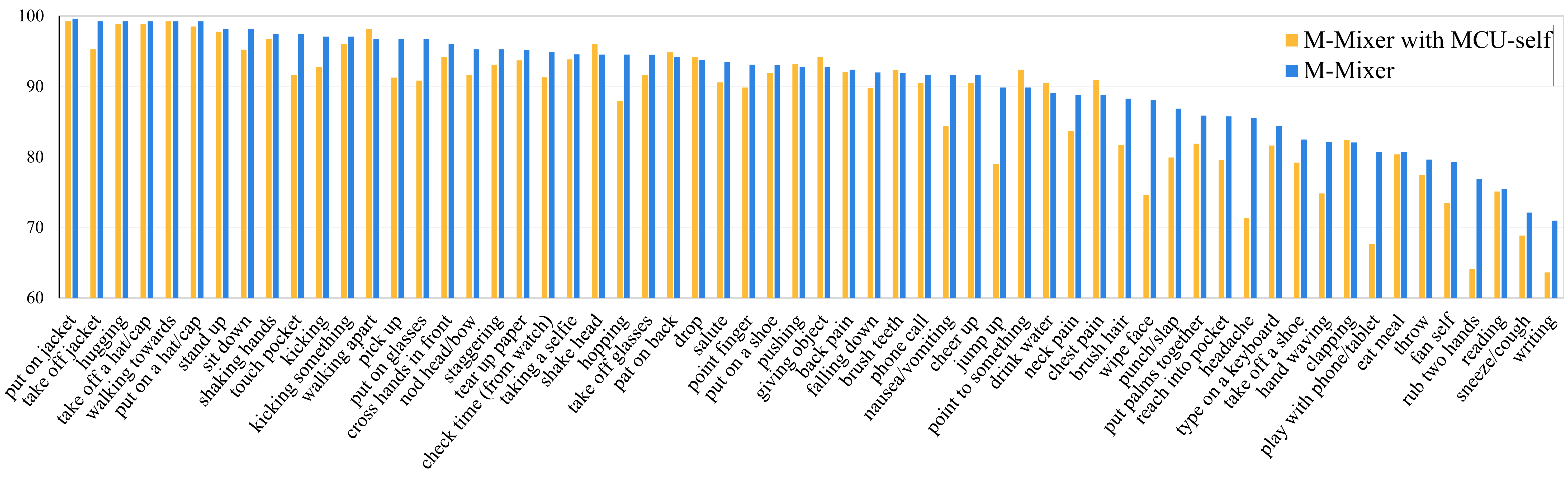}
    \caption{\textbf{Class-wise performance of M-Mixer network on NTU RGB+D 60~\cite{ntu60}.}
    60 action classes are listed in descending order according to the performance of our M-Mixer.
    Compared to the results of M-Mixer with MCU-self, the proposed method attains higher performances in most of the action classes. 
    In particular, M-Mixer network exhibits significant performance improvement in action classes where the performance of MCU-self is considerably low (\eg, `rub two
hands’, `headache’, and `writing’).
    }
    \label{fig:classwise}
\end{figure*}

\subsubsection{Cross-modality Action Content}
\label{sec:cm-action-content}
In order to assess the efficacy of the cross-modality action content, we strategically replace the cross-modality action content in MCU (see Eq.~\ref{eq:gi}) with the self-modality action content (\ie, $s^i$).
This replacement leads to a configuration called `MCU-self'.
Notably, the self-modality action content $s^i$ only comprises global action information, as it is drived from the same modality as a sequence  $f^i_{1:T}$ to be encoded.
On the other hand, the cross-modality action content $g^i$ includes not only global action information but also complementary information from other modalities.

\begin{table}[t]
\renewcommand{\arraystretch}{1.1}
\begin{center}
\begin{tabular}{c|cc|c}
\hline
\multirow{2}{*}{Modality} & \multicolumn{2}{c|}{Accuracy(\%)} & \multirow{2}{*}{$\Delta$(\%p)} \\ \cline{2-3}
 & MCU-self & MCU &  \\ \hline \hline
RGB stream&  56.61& \textbf{79.42} & + 22.81  \\
Depth stream&  84.31& \textbf{88.59} & + 4.28\\ \cdashline{1-4}
M-Mixer &  88.17& \textbf{90.77} & + 2.60 \\ \hline
\end{tabular}
\end{center}
\caption{\textbf{Experiments on the effectiveness of the cross-modality action content.} For MCU-self, we use $c^i$ instead of $g^i$ to MCU.
A single modality represents the performance of each modality stream.
$\Delta$ indicates performance differences between MCU-self and MCU.
The best scores are marked in \textbf{bold}.}
\label{tab:cross-modal}
\end{table}

Furthermore, to thoroughly examine the impact of the cross-modality action content, we evaluate the action recognition performances of individual modalities in our M-Mixer with MCU or MCU-self.
In other words, we assess the performance of RGB and depth features when utilizing MCU and MCU-self configurations.
Specifically, we train two additional fully-connected layers to classify an action class based on the final hidden state $h^i_T$, as follows:
\begin{align}
    p^i = \xi \left (\mathbf{W}_{p}^i h^i_T\right ),
\end{align}
where $p^i$ is a probability distribution of the $i$-th modality, and $\mathbf{W}_{p}^i \in \mathbb{R}^{d_h \times C}$ is a learnable matrix.
To train the two classifiers, we use a loss function $L_{self} = L^1 + L^2$. 
Here, $L^i$ is based on the standard cross-entropy loss, as follows:
\begin{align}
    L^i = \sum^C_{c=1} y_c \log \left ( p_c^i \right ),
\end{align}
where $p_c^i$ is the probability of the $k$-th action class for the $i$-th modality and $i=1,2$.
Note that the whole weights of M-Mixer network are fixed during the training of the classifiers.

Table~\ref{tab:cross-modal} presents the results of comparative experiments between MCU and MCU-self.
With RGB and depth modalities, MCU-self obtains an accuracy of 88.17\%.
Meanwhile, our MCU achieves an accuracy of 90.77\%, which is 2.60\%p higher than that of MCU-self.
These results demonstrate the effectiveness of the cross-modality action content.
 
Compared to the self-modality action content, the cross-modality action content contains complementary information from other modalities as well as global action content.
Specifically, the RGB feature is strengthened with depth information, and the depth feature is augmented by RGB information in the setting of this experiment.
As a result, our MCU achieves 79.42\% accuracy in the RGB stream and 88.59\% accuracy in the depth stream, which are 22.81\%p and 4.28\%p higher than RGB and depth streams of MCU-self, respectively.
From these results, we demonstrate that the cross-modality action content effectively provides additional information across modalities and our MCU successfully utilizes complementary information in temporal encoding.

In Fig.~\ref{fig:classwise}, we report class-wise performances of the proposed M-Mixer and M-Mixer with MCU-self.
The 60 action classes of NTU RGB+D 60~\cite{ntu60} are sorted in descending order based on the performances of our M-Mixer.
In most of the action classes, our M-Mixer achieves higher performances than using MCU-self.
Especially, the proposed M-Mixer has significant performance improvements in the action classes that have considerably lower performance in MCU-self. (\eg, `rub two hands', `headache', and `writing').

\begin{table}[t]
\begin{center}
\begin{tabular}{ccc|c}
\hline
CFEM & MCU & \begin{tabular}[c]{@{}c@{}}Multi-modal \\ feature bank\end{tabular} & Accuracy (\%) \\ \hline
 &  \checkmark&  &  90.77\\
 &  \checkmark&  \checkmark&  91.31\\
 \checkmark&  \checkmark&  &  91.41\\ \hdashline
 \checkmark & \checkmark & \checkmark & \textbf{91.94} \\  \hline
\end{tabular}
\end{center}
\vspace{0.05cm}
\caption{\textbf{Ablation study on the proposed M-Mixer network.} M-Mixer network has three important components: CFEM, MCU, and multi-modal feature bank. 
The proposed M-Mixer network incorporating both CFEM and the multi-modal feature bank achieves the highest performance.
The best score are marked in \textbf{bold}.}
\label{tab:ablation_mmixer}
\end{table}

\subsection{Ablation Studies}
In this section, we conduct extensive experiments to demonstrate the efficacy of the proposed M-Mixer network.
All experiments in this section are conducted on NTU RGB+D 60~\cite{ntu60} using RGB and depth modalities with a ResNet18 backbone.

\subsubsection{The Proposed M-Mixer network}
Our M-Mixer network has three key components: CFEM, MCU, and the multi-modal feature bank.
In order to assess the strengths and contributions of each model component, we perform ablation experiments on the three components and present the corresponding performance metrics in Table~\ref{tab:ablation_mmixer}.
In these experiments, we replace CFEM with spatio-temporal average pooling to vectorize a feature tensor, and we use simple concatenation of hidden state features instead of the multi-modal feature bank.
Compared to a baseline model~\cite{m-mixer} that only has MCU, the addition of CFEM or the multi-modal feature bank results in performance improvements of 0.54\%p and 0.64\%p.
In conclusion, by incorporating both CFEM and the multi-modal feature bank in our M-Mixer network, we achieve a significant performance improvement of 1.17\%p in comparison to the baseline model using only MCU.

\subsubsection{Complementary Feature Extraction Module (CFEM)}
\begin{table}[t]
\renewcommand{\arraystretch}{1.1}
\begin{center}
\begin{tabular}{c|c c|c}
\hline
 \multirow{2}{*}{Exp.} & \multicolumn{2}{c|}{Method} &  \multirow{2}{*}{Accuracy (\%)}\\
& Encoding block & Decoding block &  \\ \hline
\uppercase\expandafter{\romannumeral1} & Avg. pool (S) & Avg. pool(T) & 90.85\\
\uppercase\expandafter{\romannumeral2} & Avg. pool (S) & MHA & 91.18\\
\uppercase\expandafter{\romannumeral3} & Avg. pool (T) &  MHA & 91.30\\ 
\uppercase\expandafter{\romannumeral4} & Conv3d + Avg. pool (S) & Avg.pool(T) & 91.03\\
\uppercase\expandafter{\romannumeral5} & Conv3d + Avg. pool (S) & MHA & 91.68\\ \cdashline{1-4}
\uppercase\expandafter{\romannumeral6} & Conv3d + Avg. pool (T)& MHA & \textbf{91.94}\\ \hline
\end{tabular}
\end{center}
\vspace{0.05cm}
\caption{\textbf{Ablation study on Complementary Feature Extraction Module (CFEM).} 
'Avg. pool (S)' and 'Avg. pool (T)' indicates average pooling across spatial and temporal axis, respectively. 
`Conv3d' indicates two 3D convolutional layers with $2 \times 1 \times$ kernal size. 
And, `MHA' refers to a multi-head attention layer that operates with a query embedding feature.
The best score is marked in \textbf{bold}.}
\label{tab:cfem}
\end{table}
\label{sec:exp:ablation of cfem}
In this section, we conduct ablation experiments on the proposed CFEM architecture to validate its effectiveness.
The results of the ablation experiments are presented in Table~\ref{tab:cfem}. 
To investigate the impact of the encoding and decoding blocks within CFEM, we conduct experiments with various combinations of four encoding block settings and two decoding block settings
The encoding block settings include spatial average pooling and temporal average pooling with or without 3D convolutional layers, while the decoding block settings consist of temporal average pooling and a multi-head attention layer with a learnable query embedding.
We observe a marginal performance improvement when using the multi-head attention for the decoding block compared to using simple average pooling (Exp. \uppercase\expandafter{\romannumeral1} and Exp. \uppercase\expandafter{\romannumeral2}).
By incorporating a learnable query embedding, the model can effectively capture and represent the relationships between different modalities. 
Additionally, we find that the encoding block with temporal average pooling outperforms one with the spatial average pooling (Exp. \uppercase\expandafter{\romannumeral2} and Exp. \uppercase\expandafter{\romannumeral3}).
Our findings demonstrate that the cross-modality action content information is more prominently manifested in the spatial domain.

The aforementioned trend becomes even more evident when utilizing the encoding block with 3D convolutional layers.
In comparison to the results of Exp. \uppercase\expandafter{\romannumeral4} and Exp. \uppercase\expandafter{\romannumeral5}, incorporating the decoding block with multi-head attention leads to a performance improvement of 0.65\%p.
Ultimately, the combination of the encoding block with 3D convolutional layers and temporal average pooling and the decoding block with multi-head attention achieves an accuracy of 91.94\% (Exp. \uppercase\expandafter{\romannumeral6}).
Additionally, the 3D CNN layer is capable of modeling both spatial and temporal information in the data, which is not explored in the backbone network, and allowing for a comprehensive understanding of the cross-modal information. 
Overall, our proposed CFEM contributes to the extraction of relevant and discriminative features for the multi-modal action recognition.


\begin{table}[t]
\begin{center}
\begin{tabular}{c|c}
\hline
Method & Accuracy (\%) \\ \hline
Concatenation & 91.07\\
GRU~\cite{gru} & 90.77\\
Transformer & 91.72\\ \cdashline{1-2}
Multi-modal feature bank & 91.94\\ \hline
\end{tabular}
\end{center}
\vspace{0.05cm}
\caption{\textbf{Comparisons of the multi-modal feature bank with three substitutes: concatenation, GRU, and Transformer.}
For the experiments with GRU and Transformer, we use the concatenated feature of all modality-specific hidden state features as an input.
Additionally, in the case of Transformer, we utilize a class token to fuse and summarize action information across modality and time-step.
}
\label{tab:mmfb}
\end{table}
\subsubsection{Multi-modal feature bank}
\label{sec:ablation_mmmb}
To demonstrate the effectiveness of the multi-modal feature bank, we strategically substitute it with three alternatives: simple concatenation, GRU~\cite{gru}, and Transformer.
Table~\ref{tab:mmfb} shows the experimental results of M-Mixer network with the simple concatenation, GRU, Transformer, and the multi-modal feature bank.
When employing simple concatenation, M-Mixer network achieves an accuracy of 91.07\%.
In this setup, we use a concatenated feature comprising the last hidden state from all modalities to predict an action class.
In the experiment with GRU, hidden state features from all modalities are concatenated and taken as an input, resulting in an accuracy of 90.77\%.
For the experiment with Transformer, we employ two Transformer encoder layers in~\cite{vit} with 4 heads.
The input is a concatenated feature of the hidden states from all modalities along with a class token. 
This configuration achieves an accuracy of 91.72\%.
Note that the Transformer has 1,539 times more parameters compared to the proposed multi-modal feature bank (6,304.78K \textit{vs} 4.10K).
However, the multi-modal feature bank achieves the highest performance of 91.94\%.
From these results, we demonstrate the effectiveness of the multi-modal feature bank in capturing multi-modal information in a parameter-efficient manner.

\subsection{Comparisons with state-of-the-arts}
\begin{table}[t]
\renewcommand{\arraystretch}{1.1}
\begin{center}
\begin{tabular}{c|c|c|c}
\hline
Method & Backbone & Modality & Accuracy(\%) \\ \hline \hline
Sharoudy \etal~\cite{shahroudy2017deep} & - & R + D & 74.86 \\
Liu \etal~\cite{liu2018viewpoint} & - & R + D &77.5 \\
ADMD~\cite{garcia_admd} & ResNet50 & R + D & 77.74 \\
Dhiman \etal~\cite{mstd} & Incep.V3& R + D & 79.4 \\
Garcia \etal~\cite{garcia_eccv} & ResNet50& R + D & 79.73 \\
c-ConvNet~\cite{rgbd_aaai} &VGG16& R + D &86.42 \\
DMCL~\cite{garcia_dmcl}  & R(2+1)D-18 & R + D + F & 87.25 \\
Wang \etal~\cite{mmar_lstm}  & ResNet50 & R + D + F & 89.51 \\ 
ActionMAE ~\cite{actionmae}& ResNet34 & R + D & 92.5 \\
ActionMAE ~\cite{actionmae}& ResNet34 & R + D + I & 93.0 \\\cdashline{1-4}
\multirow{3}{*}{M-Mixer} & ResNet18 & R + D&  91.94\\
 & ResNet34 & R + D &  \textbf{92.54} \\
     & ResNet34 & R + D + I &  \textbf{93.16}\\\hline
\end{tabular}
\end{center}
\vspace{0.05cm}
\caption{\textbf{Performance Comparison on NTU RGB+D 60~\cite{ntu60}}. `R', `D', `F', and `I' indicate RGB, depth, optical-flow, and infrared modalities, respectively. 
To ensure a precise comparison, we present the accuracy values up to two decimal places for all papers except those that specifically indicate accuracy rounded to the first decimal place.
The best scores on each modality setting are marked in \textbf{bold}.} 
\label{tab:ntu-60}
\end{table}

We compare our M-Mixer network with state-of-the-art methods on NTU RGB+D 60~\cite{ntu60}, NTU RGB+D 120~\cite{ntu120}, and NW-UCLA~\cite{nwucla} for multi-modal action recognition.

\begin{figure*}[t!]
    \centering
    \includegraphics[width=0.95\textwidth]{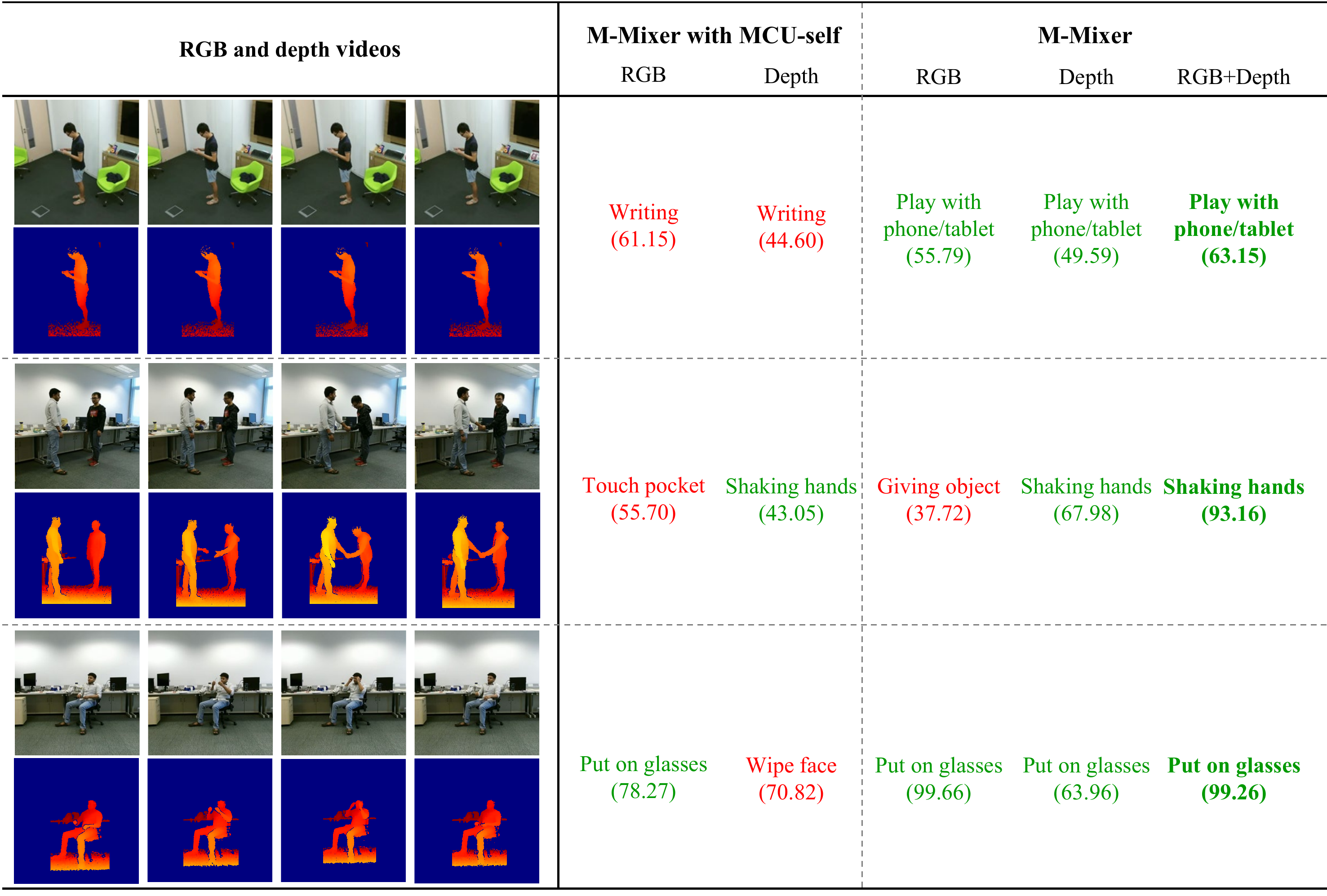}
    \caption{\textbf{Examples of the results from M-Mixer network on NTU RGB+D 60~\cite{ntu60}.} 
    Predicted results consistent with ground-truth are colored in \textcolor{green}{green}, otherwise in \textcolor{red}{red}. 
    `RGB', `Depth', and `RGB+Depth' indicate prediction results from its respective stream.
    Also, the confidence scores of predictions are presented in parentheses.
    For better visualization, the depth images are colorized following the method described in~\cite{garcia_admd}.
    }
    \label{fig:result}
\end{figure*}
\begin{table}[t]
\renewcommand{\arraystretch}{1.1}
\begin{center}
\begin{tabular}{c|c|c|c}
\hline
Method & Backbone & Modality & Accuracy(\%) \\ \hline \hline
Liu \etal~\cite{ntu120} & VGG & R + D & 61.9 \\
DMCL~\cite{garcia_dmcl} & R(2+1)D-18 & R + D + F & 89.74 \\
ActionMAE~\cite{actionmae} & ResNet34 & R + D& 91.5 \\
ActionMAE~\cite{actionmae} & ResNet34 & R + D + I& 92.3 \\\cdashline{1-4}
\multirow{2}{*}{M-Mixer} & ResNet34& R + D &  \textbf{91.54}\\
 & ResNet34 & R + D + I& \textbf{92.66} \\\hline
\end{tabular}
\end{center}
\vspace{0.05cm}
\caption{\textbf{Performance Comparison on NTU RGB+D 120~\cite{ntu120}.} `R', `D', `F', and `I' indicate RGB, depth, optical-flow, and infrared modalities, respectively. 
To ensure a precise comparison, we present the accuracy values up to two decimal places for all papers except those that specifically indicate accuracy rounded to the first decimal place.
The best scores on each modality setting are marked in \textbf{bold}.}
\label{tab:ntu-120}
\end{table}
\subsubsection{NTU RGB+D 60} 
In Table~\ref{tab:ntu-60}, we compare the performances of our M-Mixer and state-of-the-art approaches on NTU RGB+D 60.
Our M-Mixer network, which utilize RGB and depth modalities with a ResNet18 backbone, achieves an impressive accuracy of 91.94\%, suprassing DMCL~\cite{garcia_dmcl} by 4.69\%p and the method proposed by Wang \etal~\cite{mmar_lstm} by 2.43\%p.
Note that those methods incorporate additional information of optical flow.
Also, with ResNet34 backbone, our M-Mixer network achieves a performance of 92.54\% when using RGB and depth modalities, and 93.16\% when incorporating RGB, depth, and infrared modalities, surpassing the performance of ActionMAE~\cite{actionmae}.
It is noteworthy that the proposed M-Mixer consists of 54.68M parameters, while ActionMAE has 81.73M parameters, which is 49.46\% more than ours.
These findings highlight the effectiveness of our proposed M-Mixer in capturing the relationships between modalities and temporal context while achieving competitive performance with fewer parameters.

\subsubsection{NTU RGB+D 120} 
Table~\ref{tab:ntu-120} shows performance comparisons on NTU RGB+D 120.
Despite the NTU RGB+D 120 dataset containing twice the number of samples and classes compared to NTU RGB+D 60, our M-Mixer achieves state-of-the-art performance.
Using RGB and depth modalities, out M-Mixer network achieves an accuracy of 91.54\%, surprassing the method proposed by the method proposed by Liu \etal~\cite{ntu120} by 29.6\%p and DMCL~\cite{garcia_dmcl} by 1.80\%p.
This performance are comparable to ActionMAE~\cite{actionmae} with fewer parameters (54.71M \textit{vs} 81.76M).
Furthermore, when incorporating RGB, depth, and infrared modalities, our M-Mixer network achieves an accuracy 92.66\%, outperforming ActionMAE.

\begin{table}[t]
\renewcommand{\arraystretch}{1.1}
\begin{center}
\begin{tabular}{c|c|c|c}
\hline
Method & Backbone & Modality & Accuracy(\%) \\ \hline \hline
Garcia \etal~\cite{garcia_eccv} & ResNet50& R + D & 88.87 \\
ADMD~\cite{garcia_admd} & ResNet50& R + D & 89.93 \\
Dhiman \etal~\cite{mstd} & Incep.V3& R + D & 84.58 \\
DMCL~\cite{garcia_dmcl} & R(2+1)D-18 & R + D + F &  93.79\\ 
ActionMAE~\cite{actionmae} & ResNet34 & R+ D & 91.0 \\ \cdashline{1-4}
M-Mixer & ResNet18 & R + D &  \textbf{94.86}\\ \hline
\end{tabular}
\end{center}
\vspace{0.05cm}
\caption{\textbf{Performance Comparison on NW-UCLA~\cite{nwucla}.} 
`R', `D', and `F' indicate RGB, depth, and optical-flow modalities, respectively.
To ensure a precise comparison, we present the accuracy values up to two decimal places for all papers except those that specifically indicate accuracy rounded to the first decimal place.
The best scores are marked in \textbf{bold}.
}
\label{tab:nwucla}
\end{table}

\subsubsection{NW-UCLA}
In Table~\ref{tab:nwucla}, we summarize the results on NW-UCLA.
Our M-Mixer network outperforms the existing state-of-the-art methods by achieving a performance of 94.86\%.
This performance is 1.07\% higher than DMCL that utilizes additional flow modality, and 3.9\% higher than ActionMAE~\cite{actionmae} with RGB and depth.
These results convincingly demonstrate the effectiveness of our proposed M-Mixer network, particularly in the context of small-sized datasets.

\subsection{Examples of Results}
Figure~\ref{fig:result} shows the prediction results of M-Mixer on three sample videos of the NTU RGB+D 60~\cite{ntu60}.
To clearly see the efficacy of cross-modality action content, we also report the prediction results of RGB and depth streams in M-Mixer with MCU-self.
In addition, we present the confidence score of each prediction in parentheses.
We observe that the proposed M-Mixer network improves the prediction results of both RGB and depth streams in comparison to using MCU-self.
For example, in the last row in Fig.~\ref{fig:result}, the depth stream in M-Mixer with MCU-self incorrectly predicts `put on glasses' to `wipe face' due to the absence of visual appearance information.
In contrast, both the RGB and depth streams in the proposed M-Mixer network classify the video correctly to `put on glasses'.
These results show that using cross-modality action content is more effective in leveraging complementary information from other modalities than self-modality action content.


\section{Conclusion}
\label{chap:conclusion}
In this paper, we address two important factors for multi-modal action recognition: exploiting complementary information from multiple modalities and leveraging temporal context of the action.
To achieve these, we have proposed the novel network, named Modality Mixer (M-Mixer) network, which comprises a simple yet effective recurrent unit, called Multi-modal Contextualization Unit (MCU).
The proposed MCU effectively models the complementary relationships between modalities, enhancing the representation of the multi-modal sequences.
The encoded feature sequences from MCU are merged through the multi-modal feature bank, capturing multi-modal action information.
Furthermore, we have presented Complementary Feature Extraction Module (CFEM) to leverage suitable complementary information and global action content.                                            
We evaluate the performances of our M-Mixer network on NTU RGB+D 60~\cite{ntu60}, NTU RGB+D 120~\cite{ntu120}, and NW-UCLA~\cite{nwucla}.
The proposed M-Mixer network outperforms the previous state-of-the-art methods, highlighting its effectiveness in capturing and leverage multi-modal cues for accurate action recognition.
Moreover, we demonstrate the effectiveness of the M-Mixer network through comprehensive ablation studies.
\bibliographystyle{IEEEtran}
\bibliography{mybib}


 

\begin{IEEEbiography}[{\includegraphics[width=1in,height=1.25in,clip,keepaspectratio]{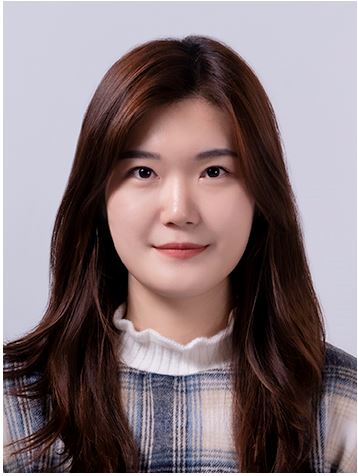}}]{Sumin Lee}
received the B.S. degree in the School of Electronic engineering from Kyungpook National University, Daegu, South Korea, in 2018, and the M.S. degree in the school of electrical engineering from Korea Advanced Institue of Science and Technology  (KAIST), Daejeon, South Korea, in 2020. She is currently pursuing the Ph.D. degree in electrical engineering with the school of electrical engineering from KAIST. Her research interest includes action recognition, anticipation, and localization for video understanding.
\end{IEEEbiography}

\begin{IEEEbiography}[{\includegraphics[width=1in,height=1.25in,clip,keepaspectratio]{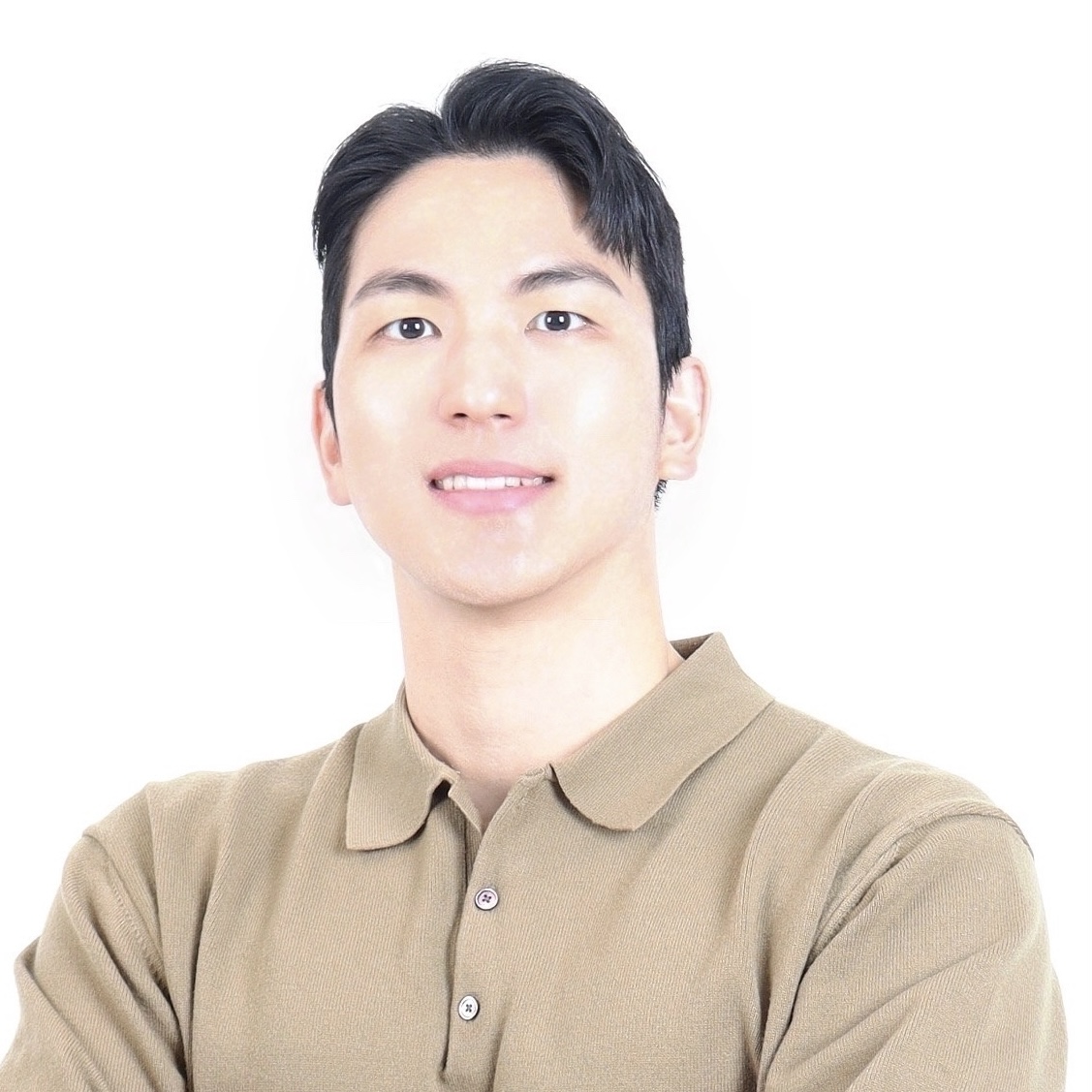}}]{Sangmin Woo} is currently pursuing a Ph.D. degree in Electrical Engineering at the Korea Advanced Institute of Science and Technology (KAIST). In 2021, he completed his M.S. degree in Electrical Engineering and Computer Science from the Gwangju Institute of Science and Technology (GIST), after obtaining his B.S. degree in Electrical Engineering from Kyungpook National University in 2019. His research interests primarily focus on computer vision and machine learning, with a particular emphasis on multi-modal learning.
\end{IEEEbiography}

\begin{IEEEbiography}[{\includegraphics[width=1in,height=1.25in,clip,keepaspectratio]{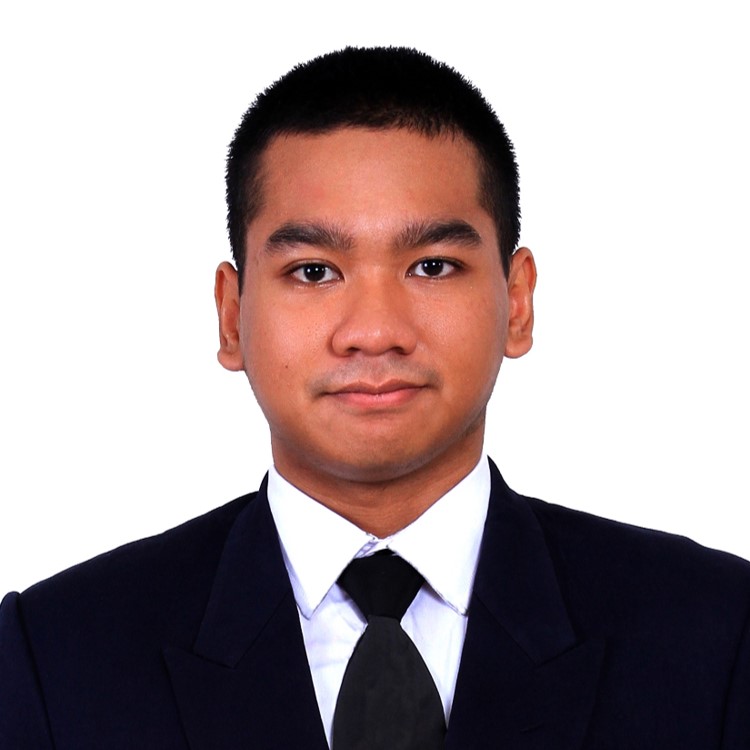}}]{Muhammad Adi Nugroho}
received the B. Eng. degree in 2016 and M. Eng. degree in 2018 from Electrical Engineering at the University of Indonesia, Jakarta, Indonesia. He is currently pursuing a Ph.D. degree in Electrical Engineering at the Korea Advanced Institute of Science and Technology (KAIST). His research interests include computer vision, action recognition,  and multi-modal learning.
\end{IEEEbiography}

\begin{IEEEbiography}[{\includegraphics[width=1in,height=1.25in,clip,keepaspectratio]{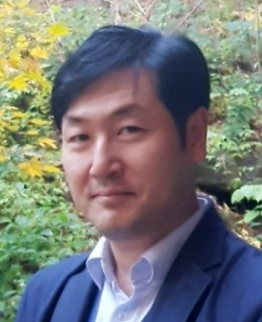}}]{Changick Kim}
received the B.S. degree in electrical engineering from Yonsei University, Seoul, South Korea, in 1989, the M.S. degree in electronics and electrical engineering from the Pohang University of Science and Technology (POSTECH), Pohang, South Korea, in 1991, and the Ph.D. degree in electrical engineering from the University of Washington, Seattle, WA, USA, in 2000. From 2000 to 2005, he was a Senior Member of Technical Staff with Epson Research and Development, Inc., Palo Alto, CA, USA. From 2005 to 2009, he was an Associate Professor with the School of Engineering, Information and Communications University, Daejeon, South Korea. Since March 2009, he has been with the School of Electrical Engineering, Korea Advanced Institute of Science and Technology (KAIST), Daejeon, Korea, where he is currently a Professor. He is also in charge of the center for security technology research, KAIST. His research interests include few shot learning, adversarial attack, and 3D reconstruction
\end{IEEEbiography}

\vfill

\end{document}